\documentclass[aps,prx,twocolumn,10pt,longbibliography,nofootinbib]{revtex4-2}

\usepackage[utf8]{inputenc}
\usepackage[T1]{fontenc}
\usepackage{microtype}
\usepackage{newtxtext,newtxmath}
\AtBeginDocument{\RenewCommandCopy\qty\SI}
\usepackage{graphicx}
\usepackage{subcaption}
\usepackage{tabularx}
\usepackage{booktabs}
\usepackage{xcolor}
\usepackage{siunitx}
\usepackage{hyperref}

\usepackage{algorithm}
\usepackage{algpseudocode}

\usepackage{amsmath,amssymb, bm}
\usepackage{physics}

\usepackage{graphicx}

\usepackage{xcolor}
\usepackage{float}
\usepackage{hyperref}
\hypersetup{
    colorlinks=true,
    linkcolor=blue,
    citecolor=blue,
    urlcolor=blue,
    pdftitle={Sculpting Quantum Landscapes},
    pdfauthor={Marwan Ait Haddou}
}

\begin{document}

\title{Sculpting Quantum Landscapes:\\ Fubini--Study Metric Conditioning for Geometry-Aware Learning in Parameterized Quantum Circuits}
\author{Marwan Ait Haddou}
\author{Mohamed Bennai}
\affiliation{Quantum Physics and Spintronic Team, LPMC, Faculty of Sciences Ben M’sick, Hassan II University of Casablanca, Morocco}
\date{July 2025}
\begin{abstract}
We introduce a novel meta-learning framework that explicitly conditions the Fubini--Study (FS) metric tensor of Parameterized Quantum Circuits (PQCs) to address the challenges of barren plateaus in Variational Quantum Algorithms (VQAs). While VQAs offer a promising paradigm for near-term quantum computing, their practical efficacy is often limited by poorly conditioned optimization landscapes. Our theoretical analysis establishes that the logarithmic condition number ($\log \kappa$) of the FS metric is a pivotal geometric quantity governing trainability, optimization dynamics, and generalization bounds. We propose \textsc{Sculpture}, a classical meta-model that learns to generate data-dependent PQC initializations by minimizing $\log \kappa(G)$, thereby promoting an isotropic and tractable parameter space.

Empirical results demonstrate that meta-training successfully reduces $\log \kappa$ from approximately 1.47 to 0.64, achieved by substantially increasing the minimum eigenvalue ($\lambda_{\min}$) and slightly decreasing the maximum eigenvalue ($\lambda_{\max}$), effectively mitigating barren plateaus. This learned conditioning robustly generalizes to unseen data, consistently yielding well-conditioned PQC initializations. In a downstream hybrid quantum-classical classification task on the Kaggle diabetes dataset, we demonstrate that increasing the meta-scaling coefficient $\lambda$ significantly enhances training efficiency, resulting in faster convergence, lower training loss, and reduced gradient norms. Crucially, higher $\lambda$ values consistently yield superior generalization performance, with test accuracy sharply increasing from around 0.68 to over 0.78. Our findings confirm that sculpting the quantum landscape via meta-learning provides a principled form of geometric regularization, fundamentally improving the trainability, optimization dynamics, and generalization capabilities of PQCs. This work offers a powerful avenue for developing more robust and efficient VQAs.
\end{abstract}

\maketitle

\section{Introduction}\label{sec:introduction}

Variational quantum algorithms (VQAs) represent one of the most promising avenues for near-term quantum computing, leveraging parameterized quantum circuits (PQCs) optimized to solve tasks ranging from quantum chemistry and materials science to diverse applications in quantum machine learning (QML), such as credit card fraud detection \cite{alami2025comparativeperformanceanalysisquantum}, brain tumor classification \cite{haddou2025hqcmebtchybridquantumclassicalmodel}, and federated adversarial learning \cite{maouaki2025qfalquantumfederatedadversarial}. However, the practical deployment of VQAs is hindered by critical challenges such as barren plateaus in the optimization landscape \cite{mcclean2018barren}, slow convergence rates, and poor generalization capabilities \cite{abbas2021power}. Addressing these issues demands a principled understanding of the underlying geometry induced by PQCs on the quantum state manifold.

The \textbf{Fubini-Study (FS) metric tensor} naturally arises as a Riemannian metric on the projective Hilbert space, capturing the intrinsic geometry of the quantum states generated by a PQC. By quantifying infinitesimal distinguishability between parameter-dependent quantum states, the FS metric encapsulates fundamental information about parameter sensitivity, expressiveness, and curvature of the model manifold. Despite its foundational importance, a comprehensive theoretical framework linking FS metric spectral properties to optimization dynamics and generalization in VQAs remains underdeveloped.

In this work, we present a rigorous geometric analysis emphasizing the \emph{spectral conditioning} of the FS metric tensor as a pivotal factor governing trainability and generalization in PQCs. Specifically, we investigate the logarithmic condition number \(\log \kappa(G(\theta, x))\) of the FS metric tensor evaluated \emph{locally} at a parameter point \(\theta\) and for a given input data point \(x\). This local condition number measures anisotropy in parameter sensitivity through the ratio of largest to smallest eigenvalues of \(G(\theta, x)\). Poor conditioning leads to ill-scaled gradients and slow convergence, while improved conditioning promotes isotropic sensitivity and more efficient optimization.

Throughout this paper, whenever we refer to \(\log \kappa\), we implicitly mean this local quantity \(\log \kappa(G(\theta, x))\), emphasizing that it is a dynamic, data- and parameter-dependent measure rather than a fixed global property of the PQC.

Building on this insight, we introduce a composite parameterization scheme where PQC parameters are decomposed into task-specific parameters and meta-learned data-dependent corrections aimed at minimizing \(\log \kappa(G)\). We theoretically predict that modulating the influence of meta-learned parameters improves spectral conditioning, thereby enhancing convergence speed and tightening PAC-Bayes generalization bounds. Furthermore, we highlight how this geometric regularization synergizes with standard gradient descent optimization and could be further exploited by natural gradient methods.

Our contributions offer a unifying geometric perspective that connects the intrinsic curvature of PQCs to their empirical performance, providing predictive theoretical tools for designing more trainable and generalizable quantum learning models. This work lays the foundation for future experimental validation and algorithmic innovation in quantum machine learning.

The remainder of this paper is organized as follows: Section \ref{sec:theoretical_framework} presents the theoretical framework, detailing the Fubini--Study metric and its spectral properties, and introducing the composite parameterization scheme. Section \ref{sec:experiments} describes our experimental setup, including the PQC architecture, the meta-model design, and the meta-training procedure. Section \ref{sec:analysis_trainability} then analyzes the evolution of trainability metrics during meta-training and the characteristics of the learned parameter distributions. Finally, Section \ref{sec:downstream_results} presents the results of the downstream classification task, demonstrating the impact of $\lambda$-scaling on optimization efficiency and generalization performance.
\section{Related Work}\label{sec:related work}
Variational Quantum Algorithms (VQAs) have emerged as a leading paradigm for leveraging near-term quantum computers across diverse applications, from quantum chemistry to machine learning \cite{cerezo2021variational}. However, their practical efficacy is often hindered by significant challenges, notably the presence of barren plateaus in the optimization landscape \cite{mcclean2018barren}. As highlighted by Gelman (2024) in a recent survey, barren plateaus are a formidable obstacle for hybrid quantum-classical algorithms, leading to flat loss function landscapes that impede gradient-based optimization due to vanishing gradients \cite{gelman2024survey}. This phenomenon shares parallels with vanishing gradient issues in classical neural networks arising from large parameter spaces and non-convex landscapes \cite{gelman2024survey}.

Efforts to mitigate barren plateaus have explored various strategies, often targeting their underlying causes. These include careful initialization schemes to avoid regions of vanishing gradients \cite{grant2019initialization}, specific circuit architectures designed to enhance trainability and avoid deep, unstructured ansatzes prone to plateaus \cite{wang2021noise, gelman2024survey}, and the use of local cost functions instead of global ones to maintain gradient magnitudes \cite{cerezo2021cost, gelman2024survey}. Other factors contributing to barren plateaus, such as information scrambling, noise, and highly entangled circuits, have also been identified, prompting research into noise-resilient designs and entanglement-aware circuit construction \cite{gelman2024survey}. While effective in certain contexts, these methods often provide indirect control over the global landscape geometry.

The Fubini--Study (FS) metric tensor naturally quantifies the intrinsic geometry of the projective Hilbert space, providing insights into parameter sensitivity and the curvature of the model manifold \cite{braunstein1994quantum}. Prior work has explored the role of quantum geometry in VQAs, with some studies focusing on the relationship between the FS metric and gradient magnitudes \cite{stokes2020quantum}, or its use in natural gradient methods to accelerate optimization \cite{amari1998natural, yamamoto2019natural}. Recent advancements have further contributed to this area, including efficient estimation techniques for the Quantum Fisher Information (proportional to the FS metric) via Stein's identity \cite{halla2025estimation}, and the development of quantum natural gradient methods incorporating geodesic corrections for improved optimization trajectories \cite{halla2025quantum}. The influence of \textit{circuit geometries} on the emergence of barren plateaus has also been a subject of investigation \cite{gelman2024survey}. While these approaches acknowledge the importance of geometry, a comprehensive framework for actively conditioning the FS metric to improve VQA performance, particularly concerning its spectral properties, remains an active area of research.

Meta-learning, or "learning to learn," has gained traction in classical machine learning for its ability to train models that can quickly adapt to new tasks or improve optimization processes \cite{hospedales2020meta}. More recently, meta-learning concepts have been applied to quantum computing, for instance, to learn optimal initializations for VQAs \cite{liu2021meta} or to design quantum neural networks \cite{schuld2020circuit}. Our work distinguishes itself by explicitly leveraging meta-learning to directly optimize the spectral conditioning of the Fubini--Study metric, providing a principled, geometry-aware approach to enhance VQA trainability and generalization that complements existing strategies. This direct focus on the FS metric's condition number offers a novel perspective compared to prior meta-learning applications in quantum contexts and the general barren plateau mitigation techniques surveyed in the literature.
\section{Theoretical Framework}\label{sec:theoretical_framework}

This study is grounded in a geometric perspective on variational quantum algorithms (VQAs), in which the parameterized quantum circuit (PQC) induces a differentiable manifold of quantum states. We adopt the \textbf{Fubini--Study (FS) metric tensor} as the central construct for analyzing the structure of this manifold. Defined on the projective Hilbert space \(\mathcal{P}(\mathcal{H})\), the FS metric quantifies infinitesimal distinguishability between quantum states under parameter shifts. Mathematically, it yields a Riemannian metric \(G(\bm{\theta})\) on the state manifold \(\mathcal{M} = \{ \ket{\psi(\bm{\theta})} \}\), given by the covariance of Heisenberg-evolved generators (see Appendix Eq.~\ref{eq:fs_metric_cov}). This tensor governs both the local curvature of the model class and the parameter sensitivity, making it a foundational object for characterizing trainability, expressiveness, and generalization in quantum learning models \citep{braunstein1994quantum, amari2016information, mitarai2018quantum}.

Our analysis proceeds from the recognition that the FS metric generalizes the classical Fisher information metric to quantum systems and thereby directly informs both optimization dynamics (via natural gradient descent) and generalization bounds (via PAC-Bayes theory). In particular, we focus on the \textbf{logarithmic condition number} \(\log \kappa(G)\), which measures the anisotropy of \(G(\bm{\theta})\) through the ratio of its largest to smallest eigenvalues (Appendix Eq.~\ref{eq:condition_number}). A large \(\kappa(G)\) signals poorly conditioned parameter directions, leading to ill-scaled gradients, barren plateaus, and slow or unstable convergence \citep{mcclean2018barren, cerezo2021cost}. Conversely, minimizing \(\log \kappa(G)\) induces geometric isotropy, ensuring uniformly sensitive parameters and efficient learning dynamics. This conditioning also increases the volume element \(\sqrt{\det G}\) of the induced manifold, reflecting enhanced model capacity and expressiveness (Appendix Sec.~\ref{app:volume_element}) \citep{abbas2021power, du2021expressive}.

To operationalize this geometric insight, we consider a \textbf{composite parameterization} of the PQC:
\[
\bm{\theta}(x) = \bm{\theta}_{\mathrm{task}} + \lambda \cdot \bm{\theta}_{\mathrm{meta}}(x),
\]
where \(\bm{\theta}_{\mathrm{meta}}(x)\) is generated by a meta-model trained (or hypothesized) to minimize \(\log \kappa(G)\) across data instances \(x\) (Appendix Sec.~\ref{app:meta_model_training}). The scalar \(\lambda\) modulates the meta-learned contribution, allowing us to investigate how increasing meta-guidance influences spectral conditioning.

\paragraph{Theoretical Predictions:}

\begin{itemize}
\item Increasing \(\lambda\) is predicted to decrease \(\log \kappa(G)\), thereby flattening the eigenvalue spectrum of \(G\) and promoting spectral entropy and effective rank. These spectral statistics—including the entropy
\[
\mathcal{H}(G) = -\sum_i p_i \log p_i
\]
and effective dimension
\[
d_{\mathrm{eff}} = \frac{1}{\sum_i p_i^2},
\]
where \(p_i = \lambda_i / \sum_j \lambda_j\) are the normalized eigenvalues—formally quantify the degree of isotropy in parameter sensitivity and the uniformity of information distribution across the model (see Appendix Sec.~\ref{app:spectral_stats}) \citep{abbas2021power, wu2021towards}.

\item Leveraging recent PAC-Bayes bounds for quantum models, the generalization error is theoretically controlled by the ratio \(\mathrm{Tr}(G)/\lambda_{\min}(G)\), which is bounded above by \(d \cdot \kappa(G)\). Thus, minimizing the log condition number directly tightens these bounds by reducing effective model complexity in information-theoretic terms (Appendix Sec.~\ref{app:pac_bayes_derivation}, Eq.~\ref{eq:fs_pac_bayes_surrogate}). Hence, improved FS metric conditioning is predicted to serve as a geometric regularizer that aligns the curvature of the parameter manifold with enhanced generalization performance \citep{abbas2021power, zhao2022generalization}.

\item From an optimization perspective, natural gradient descent, which adjusts gradients by \(G^{-1}\), benefits directly from well-conditioned metrics: a lower \(\kappa(G)\) guarantees more isotropic step sizes and faster convergence. The contraction factor for gradient descent in quadratic basins,
\[
\rho^2 = \left(\frac{\kappa - 1}{\kappa + 1}\right)^2,
\]
quantitatively underscores the central role of FS conditioning (Appendix Sec.~\ref{app:optimization_analysis}, Eq.~\ref{eq:grad_descent_convergence}) \citep{amari1998natural, stokes2020quantum}. Consequently, \(\lambda\)-scaling is predicted to act as a meta-learning mechanism that sculpts the geometry of the parameter space to optimize both convergence speed and generalization.
\end{itemize}

In sum, this theoretical framework links the spectral properties of the FS metric tensor to the optimization dynamics and generalization capacity of PQCs, framing \(\log \kappa(G)\) as a unifying geometric quantity. By controlling FS anisotropy through data-dependent parameterizations, we posit a principled form of geometric regularization that can underpin improvements in training efficiency and robustness. Experimental validation of these predictions is left for subsequent sections.

\section{Experiments}\label{sec:experiments}

\subsection{Parameterized Quantum Circuit Architecture}\label{sec:pqc_architecture}

The parameterized quantum circuit (PQC) used throughout our experiments follows a hardware-efficient ansatz design, chosen for its scalability and empirical success in near-term quantum learning tasks. The circuit operates over $N = 8$ qubits and consists of $L = 3$ layers of parametrized single-qubit rotations and entangling gates.

The PQC implements the following structure:
\begin{itemize}
    \item \textbf{Data Encoding:} Each input feature is mapped to a single qubit using a pair of rotation gates: an RX rotation for the raw value and a scaled RZ rotation to inject nonlinear feature variation. Specifically, for input vector $x \in \mathbb{R}^d$, we apply
    \[
    \texttt{RX}(x_i),\quad \texttt{RZ}(0.01 \cdot x_i), \quad \text{on qubit } i \bmod N.
    \]
    \item \textbf{Parameterized Layers:} Each layer consists of a sequence of parameterized single-qubit gates:
    \[
    \texttt{RX}(\theta), \quad \texttt{RY}(\theta), \quad \texttt{RZ}(\theta) \quad \text{on each qubit},
    \]
    followed by circular entangling gates that apply $\texttt{CNOT}$ operations between each pair of adjacent qubits and wrap around from the last to the first qubit to ensure full entanglement connectivity.
    \item \textbf{Trainable Parameters:} The total number of parameters in the PQC is
    \[
    p = N \times L \times 3 = 8 \times 3 \times 3 = 72.
    \]
    These parameters are either learned directly during task training (for $\bm{\theta}_{\text{task}}$) or generated by the meta-model (for $\bm{\theta}_{\text{meta}}(x)$).
\end{itemize}

This architecture balances expressiveness with hardware feasibility. The repeated RX-RY-RZ blocks allow the circuit to represent a rich family of quantum functions, while the nearest-neighbor plus circular entanglement pattern ensures sufficient entanglement depth without requiring long-range gates.

\noindent The ansatz is implemented using PennyLane's circuit construction primitives.

\noindent This circuit topology is well-suited for analyzing the geometric structure of the Fubini--Study metric tensor and the impact of spectral conditioning.

\subsection{Meta-Model Architecture and Training Objective}\label{sec:meta_model_arch}

The meta-learning framework centers on a classical neural network we call \textsc{Sculpture}, designed to generate data-dependent initial parameters $\bm{\theta}(x) \in \mathbb{R}^{p}$ for parameterized quantum circuits (PQCs). Given a classical input vector $x \in \mathbb{R}^{d_{\text{in}}}$, the model outputs a length-$p$ parameter vector via a shared encoder followed by multiple output heads—one per PQC parameter:
\[
\bm{\theta}(x) = \pi \cdot \sigma\left( \left[ h_1(x), h_2(x), \dots, h_p(x) \right] \right),
\]
where $\sigma(\cdot)$ is the sigmoid activation function and each $h_i$ maps from the shared hidden representation to a scalar. This architecture introduces inductive structure into the PQC parameter space, enabling input-conditioned parameter initialization.

Training the meta-model involves minimizing the logarithmic condition number $\log \kappa$ of the Fubini--Study (FS) metric tensor computed for the PQC initialized with $\bm{\theta}(x)$. The FS metric $G(\bm{\theta})$ characterizes the local Riemannian geometry of the quantum state manifold with respect to circuit parameters. A smaller condition number $\kappa = \lambda_{\max} / \lambda_{\min}$ reflects a more isotropic and trainable landscape, mitigating barren plateaus and enabling better gradient flow.

Each meta-training update proceeds as follows:
\begin{enumerate}
    \item Sample a representative input $x_{\text{rep}}$ and generate $\bm{\theta}(x_{\text{rep}})$ using the meta-model.
    \item Evaluate the average FS metric $\bar{G}$ over a batch $\{x_j\}_{j=1}^B$ of inputs with PQCs initialized at $\bm{\theta}(x_{\text{rep}})$.
    \item Compute the meta-loss $\mathcal{L}_\text{meta} = \log \kappa(\bar{G})$ from the eigenvalues of $\bar{G}$ and backpropagate through the meta-model.
    \item Update meta-model parameters using the AdamW optimizer.
\end{enumerate}
The FS metric is estimated via a block-diagonal approximation. To maintain numerical stability, eigenvalues below a threshold $\varepsilon$ are discarded. Training is monitored for degeneracies that may signal barren plateaus. This fully differentiable meta-objective enables the meta-model to learn parameter initializations explicitly optimized for circuit geometry and downstream trainability.

\subsection{Meta-Training Procedure}\label{sec:meta_training_procedure}

We now detail the meta-training loop for optimizing the parameter-generating model \textsc{Sculpture}, which learns to produce PQC parameters that yield well-conditioned Fubini--Study (FS) metrics across data instances. The meta-objective minimizes the average $\log \kappa(G)$, where $G$ is estimated via a block-diagonal approximation of the FS metric tensor across mini-batches of data.

\begin{figure}[H]
\centering
\begin{minipage}{0.95\linewidth}
\begin{flushleft}
\textbf{Algorithm 1:} Meta-Training of \textsc{Sculpture} via FS Metric Conditioning
\end{flushleft}
\vspace{-1mm}
\hrule
\vspace{1mm}
\begin{flushleft}
\textbf{Input:} Input dataset $\mathcal{D} = \{x_i\}_{i=1}^N$, PQC architecture with $p$ parameters, batch size $B$, meta-model $\mathcal{M}_\phi(x)$, optimizer $\mathcal{O}$ \\
\textbf{Output:} Meta-learned parameter generator $\mathcal{M}_\phi(x)$ that minimizes FS log-condition number
\end{flushleft}
\vspace{-1mm}
\hrule
\vspace{1mm}
\begin{flushleft}
\begin{enumerate}
    \item[\textbf{1.}] Initialize model parameters $\phi \sim \mathcal{N}(0, \sigma^2)$, set learning rate $\eta$
    \item[\textbf{2.}] \textbf{for} each meta-training step $t = 1, \dots, T$ \textbf{do}
    \begin{enumerate}
        \item[(a)] Sample a representative input $x_\text{rep} \sim \mathcal{D}$
        \item[(b)] Generate PQC parameters: $\bm{\theta}_{\text{rep}} \gets \mathcal{M}_\phi(x_\text{rep})$
        \item[(c)] Sample a batch $\{x_j\}_{j=1}^B \subset \mathcal{D}$ for estimating FS metric
        \item[(d)] For each $x_j$, construct circuit $U_{\bm{\theta}_{\text{rep}}}(x_j)$ and compute block-diagonal FS metric $G(x_j)$
        \item[(e)] Average FS metric over batch: $\bar{G} \gets \frac{1}{B} \sum_{j=1}^B G(x_j)$
        \item[(f)] Compute loss: $\mathcal{L} \gets \log \kappa(\bar{G}) = \log \left( \frac{\lambda_{\max}(\bar{G})}{\lambda_{\min}(\bar{G})} \right)$
        \item[(g)] Update $\phi \gets \phi - \eta \nabla_\phi \mathcal{L}$ via AdamW
    \end{enumerate}
    \item[\textbf{3.}] \textbf{end for}
\end{enumerate}
\end{flushleft}
\vspace{-2mm}
\hrule
\end{minipage}
\end{figure}

The meta-model is optimized using backpropagation through a differentiable estimator of $\log \kappa(G)$. The metric tensor is computed with PennyLane using the parameter-shift rule and averaged across a mini-batch of circuits initialized with shared parameters but evaluated on diverse inputs. To detect barren plateaus, we also monitor the eigenvalue spectrum of $G$ and flag degeneracies when more than 50\% of eigenvalues fall below a small threshold $\varepsilon$.

This algorithm enables the meta-learner to discover an initialization function $\mathcal{M}_\phi(x)$ that geometrically regularizes the PQC via FS conditioning, thereby enhancing convergence and generalization in downstream tasks.

\subsection{Analysis of Meta-Learned Quantum Circuit Trainability}\label{sec:analysis_trainability}

We analyze how the meta-learning framework enhances the trainability of quantum neural networks (QNNs) by examining key FS metric-derived quantities that reflect the curvature and sensitivity of the parameter space. The following subsections report the evolution of these metrics during meta-training and characterize the distributions of the learned parameters.

\subsection{Evolution of Trainability Metrics During Meta-Training}\label{sec:evolution_metrics}

\begin{figure}[H]
    \centering
    \includegraphics[width=\columnwidth]{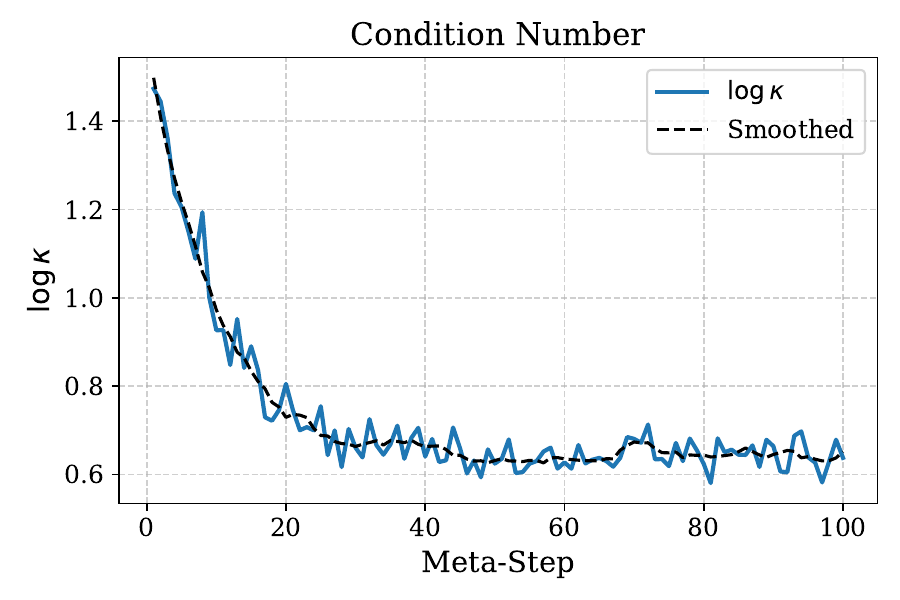}
    \caption{Logarithmic Condition Number ($\log \kappa$) of the FS metric during meta-training.}
    \label{fig:log_kappa}
\end{figure}

\begin{figure}[H]
    \centering
    \includegraphics[width=\columnwidth]{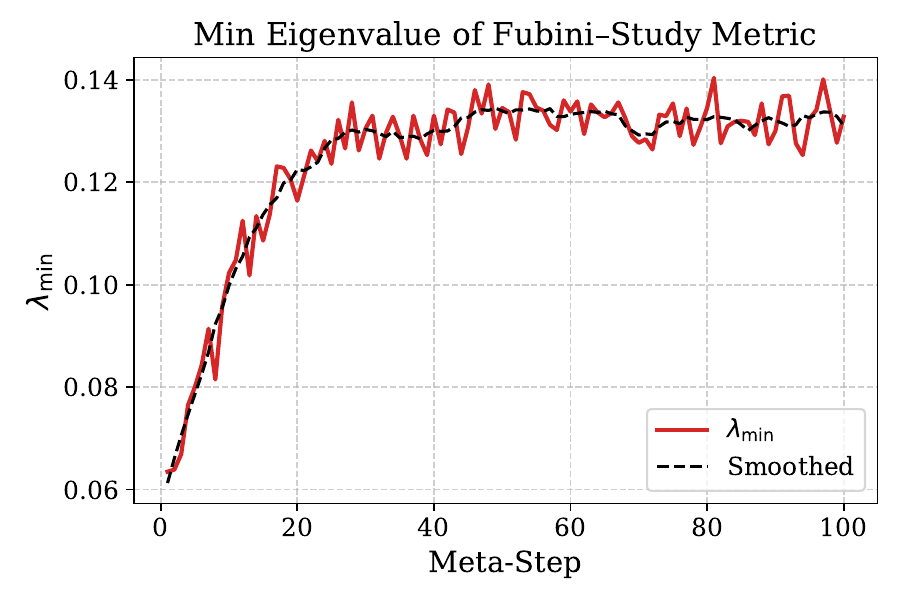}
    \caption{Minimum Eigenvalue ($\lambda_{\min}$) of the FS metric during meta-training.}
    \label{fig:lambda_min}
\end{figure}

\begin{figure}[H]
    \centering
    \includegraphics[width=\columnwidth]{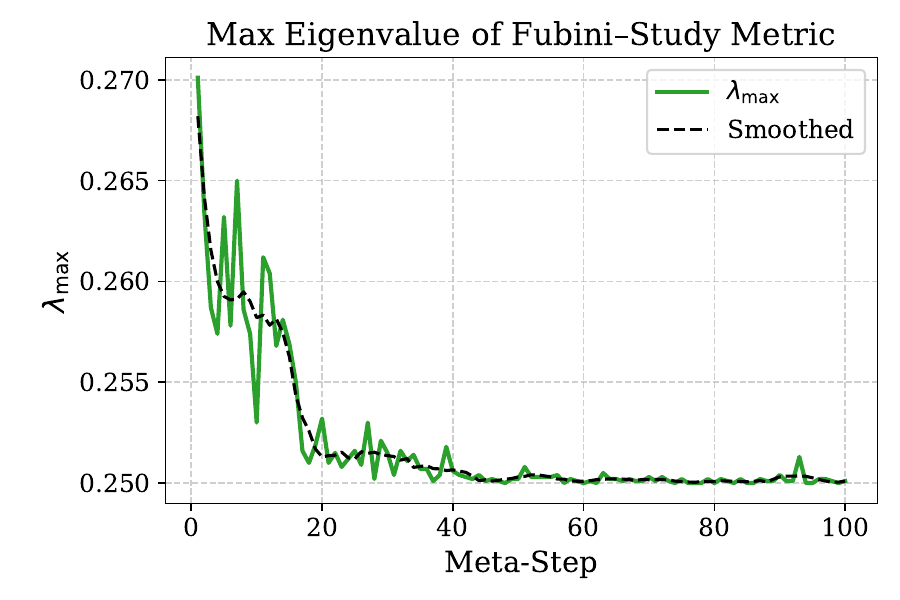}
    \caption{Maximum Eigenvalue ($\lambda_{\max}$) of the FS metric during meta-training.}
    \label{fig:lambda_max}
\end{figure}

\begin{figure}[H]
    \centering
    \includegraphics[width=\columnwidth]{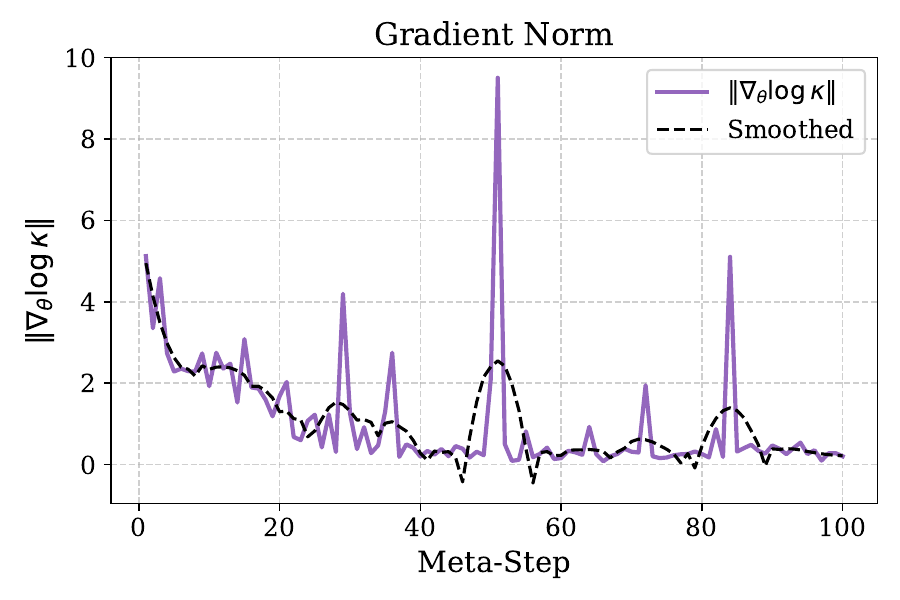}
    \caption{Gradient norm ($\|\nabla_\theta \log \kappa\|$) of the meta-model parameters during meta-training.}
    \label{fig:grad_norm}
\end{figure}

Figures~\ref{fig:log_kappa}--\ref{fig:grad_norm} illustrate the progression of key FS metric properties and the meta-learner’s gradient norm over 100 meta-training steps.

At Meta-Step 1, the logarithmic condition number $\log \kappa$ is approximately \num{1.4737}, indicating a poorly conditioned parameter space typical of random initializations prone to barren plateaus. Within the first 20–30 meta-steps, $\log \kappa$ rapidly decreases, stabilizing around \num{0.7}, and reaches \num{0.6375} by Meta-Step 100. The minimal observed value is \num{0.5804} at Meta-Step 81. This substantial reduction demonstrates that \textsc{Sculpture} effectively learns to produce PQC parameters residing in better-conditioned regions of the parameter landscape, mitigating barren plateaus and enhancing trainability. The plateauing of $\log \kappa$ toward the end indicates convergence of the meta-learning process.

Examining individual eigenvalues, $\lambda_{\min}$ and $\lambda_{\max}$ (Figures~\ref{fig:lambda_min} and~\ref{fig:lambda_max}), provides additional insight. Initially, $\lambda_{\min}$ is small (\num{0.0635}), a primary contributor to the large condition number. Through training, $\lambda_{\min}$ steadily increases, reaching about \num{0.1328} at Meta-Step 100, peaking at \num{0.1404} around Step 81. This increase implies enhanced sensitivity across a wider range of parameter directions, alleviating flat-loss landscapes associated with barren plateaus. Meanwhile, $\lambda_{\max}$, initially \num{0.2701}, steadily decreases to roughly \num{0.2501}, reflecting a slight contraction of the most sensitive directions. Together, these trends reduce $\kappa = \lambda_{\max}/\lambda_{\min}$ and promote a more isotropic FS metric.

Figure~\ref{fig:grad_norm} tracks the Euclidean norm of the meta-model parameter gradients, indicating the magnitude of updates. The norm begins high (\num{5.1212}) coinciding with steep early improvements in $\log \kappa$. As training progresses, gradient norms decrease, signaling convergence to an optimal parameter generation strategy. Minor fluctuations are observed (notably near Steps 50–60 and 82), but the overall trend is downward, ending near \num{0.2052}.

\subsection{Generalization and Learned Parameter Distributions}\label{sec:generalization_param_dist}

\begin{figure}[h!]
    \centering
    \includegraphics[width=\columnwidth]{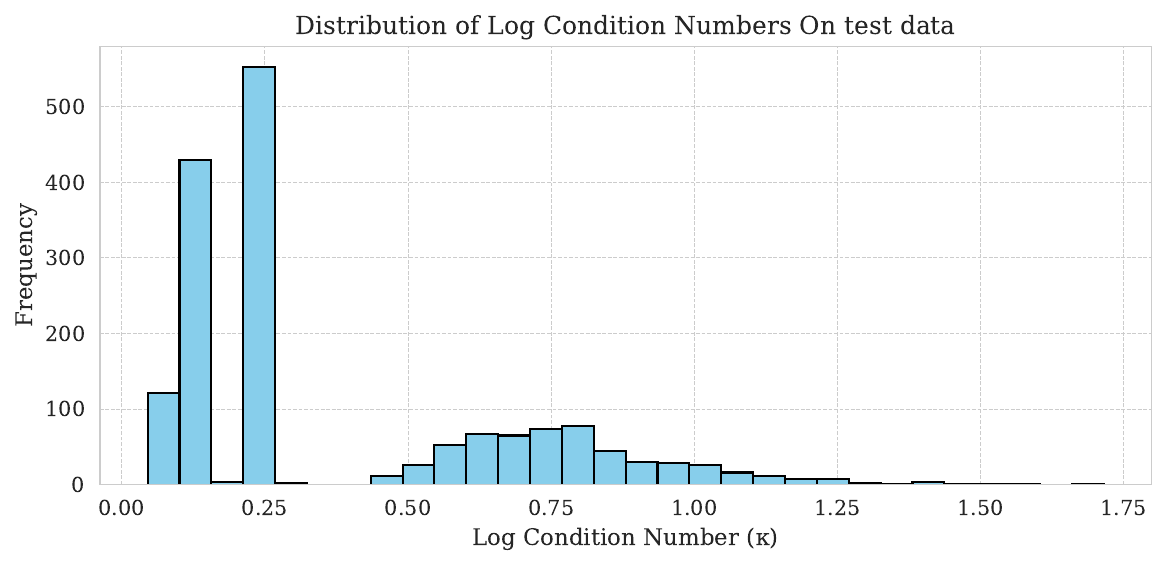}
    \caption{Distribution of $\log \kappa$ on unseen test inputs.}
    \label{fig:log_kappa_test_dist}
\end{figure}

\begin{figure}[h!]
    \centering
    \includegraphics[width=\columnwidth]{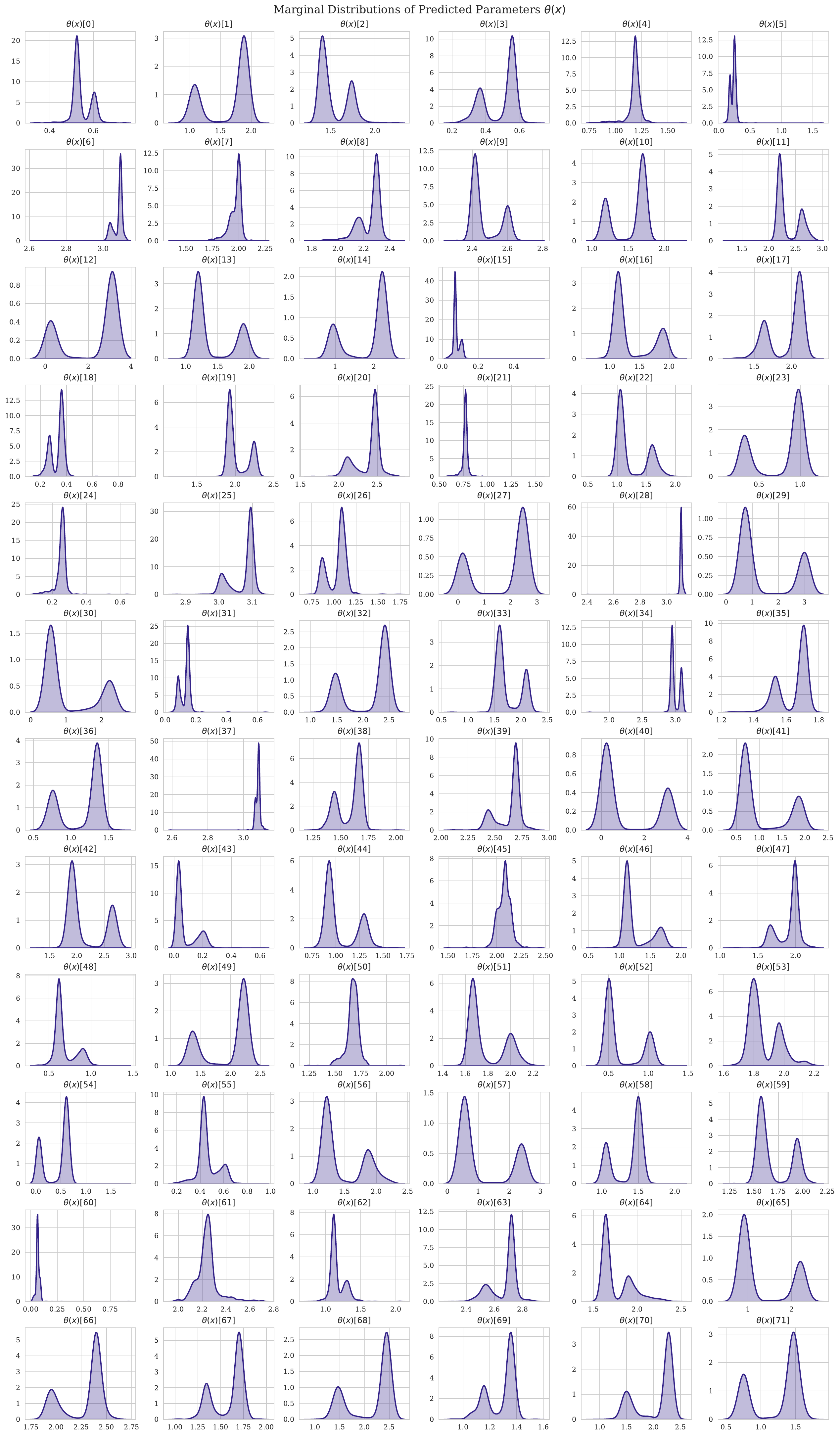}
    \caption{Marginal distributions of predicted PQC parameters $\theta_i(x)$ across test inputs.}
    \label{fig:theta_x_dist}
\end{figure}

Figure~\ref{fig:log_kappa_test_dist} depicts the distribution of $\log \kappa$ for unseen test data. The histogram reveals a strong clustering near zero, with a majority of samples exhibiting $\log \kappa < 0.25$. This suggests robust generalization of the meta-learner, which consistently produces well-conditioned PQC initializations across diverse inputs—a critical property for practical quantum learning applications. In contrast, random initializations yield broader, less favorable $\log \kappa$ distributions linked to barren plateaus.

Figure~\ref{fig:theta_x_dist} shows marginal distributions of individual predicted parameters $\theta_i(x)$ over the test set. Many exhibit multimodal structure, often with two or more distinct peaks. This multimodality suggests the meta-learner identifies multiple preferred initialization regimes within parameter space, flexibly adapting outputs to input features. This input-dependent diversity likely leverages PQC symmetries and enhances trainability by avoiding uniform, suboptimal initializations.

Moreover, parameter values concentrate within bounded intervals (e.g., $[0, \pi]$), consistent with the effects of sigmoid activations and scaling. This demonstrates the meta-learner encodes meaningful inductive biases aligned with the quantum circuit’s geometry, rather than producing arbitrary or noisy outputs.

This section establishes that the meta-model learns a data-dependent initialization strategy that substantially improves the spectral conditioning of the FS metric, thereby providing a principled means to mitigate barren plateaus and enhance PQC trainability. Subsequent sections will explore how this initialization influences downstream task performance and generalization.

\subsection{Downstream Phase: Hybrid Quantum-Classical Classification on Kaggle Diabetes Dataset}\label{sec:downstream_phase}

For the downstream evaluation, we employ a hybrid quantum-classical classifier designed to leverage the meta-learned parameterizations obtained during the meta-training phase. The classification task is performed on the Kaggle diabetes dataset, which consists of clinical feature vectors labeled for diabetes onset, serving as a standard benchmark for binary classification.

The model architecture integrates a parameterized quantum circuit (PQC) with a classical linear readout layer. Importantly, the PQC employed here is the same hardware-efficient ansatz used during meta-model training, comprising multiple layers of parameterized single-qubit rotations and entangling gates, implemented on a noiseless quantum simulator backend. Classical input features are encoded into quantum states, and the PQC outputs expectation values of Pauli-Z measurements on each qubit.

A key aspect of this model is its composite parameterization of the PQC parameters: a trainable base parameter vector augmented by a data-dependent correction generated by the frozen meta-model. This correction is scaled by a tunable scalar \(\lambda \in [0,1]\), controlling the influence of the meta-learned parameters in the downstream task. By systematically varying \(\lambda\), we investigate how meta-parameter scaling affects convergence dynamics and classification performance.

Training is performed using minibatch gradient descent with the AdamW optimizer and cross-entropy loss. To ensure stable parameter updates, gradient clipping is applied to the quantum circuit parameters. For each value of \(\lambda\), the model is trained independently over 20 epochs, with evaluation on a held-out test set to assess classification accuracy.

This protocol, sweeping \(\lambda\) over a dense range of values, enables a fine-grained analysis of the interplay between meta-learned parameter contributions and downstream task generalization. Our results confirm that increasing \(\lambda\) improves both convergence speed and test accuracy, consistent with theoretical predictions that conditioning the Fubini–Study metric via meta-learning induces beneficial geometric regularization.

Model checkpoints are saved for each training run to support reproducibility and further analysis.

\begin{algorithm}[H]
\caption{Downstream Hybrid Quantum-Classical Classifier Training with Meta-Parameter Scaling}
\label{alg:downstream_training}
\begin{algorithmic}[1]
\Require Meta-model \( \mathcal{M} \) (frozen), training data \( \mathcal{D}_{\text{train}} = \{(x_i, y_i)\} \), test data \( \mathcal{D}_{\text{test}} \), PQC ansatz \( U(\cdot) \), scaling parameters \( \{\lambda_j\}_{j=1}^L \), learning rate \(\eta\), epochs \(E\)
\Ensure Trained hybrid classifier models \(\{ \theta^{(\lambda_j)} \}\) and evaluation metrics

\For{each \(\lambda_j\) in \(\{\lambda_1, \lambda_2, \ldots, \lambda_L\}\)} 
    \State Initialize trainable PQC parameters \(\theta_{\mathrm{task}}^{(0)}\) randomly
    \For{epoch \(e = 1\) to \(E\)}
        \For{each minibatch \(B = \{(x_b, y_b)\} \subset \mathcal{D}_{\text{train}}\)}
            \State Compute meta-parameters \(\theta_{\mathrm{meta}}(x_b) = \mathcal{M}(x_b)\)
            \State Compose full PQC parameters:
            \[
                \theta = \theta_{\mathrm{task}} + \lambda_j \cdot \theta_{\mathrm{meta}}(x_b)
            \]
            \State Encode \(x_b\) into quantum states and apply PQC \(U(\theta)\)
            \State Measure observables, obtaining outputs \(o_b\)
            \State Compute classifier outputs via classical readout \(h(o_b)\)
            \State Calculate loss \(\mathcal{L}(h(o_b), y_b)\)
            \State Backpropagate gradients w.r.t. \(\theta_{\mathrm{task}}\) and update parameters:
            \[
            \theta_{\mathrm{task}} \leftarrow \theta_{\mathrm{task}} - \eta \nabla_{\theta_{\mathrm{task}}} \mathcal{L}
            \]
        \EndFor
        \State Evaluate model on \(\mathcal{D}_{\text{test}}\), record accuracy and loss
    \EndFor
    \State Save trained model \(\theta_{\mathrm{task}}^{(\lambda_j)}\) and evaluation metrics
\EndFor
\end{algorithmic}
\end{algorithm}

\section{Analysis of Downstream Results: Impact of $\lambda$-Scaling}\label{sec:downstream_results} 

The empirical results from numerical simulations, presented through both 2D line plots and 2D heatmaps, provide strong evidence for the beneficial impact of the meta-scaling coefficient $\lambda$ on the training dynamics and generalization performance of the parameterized quantum circuit (PQC) model. These findings align with the theoretical framework that posits FS metric conditioning as a form of geometric regularization.

\subsection{Optimization Efficiency: Loss and Gradient Norm Dynamics}\label{sec:optimization_efficiency}

\begin{figure}[h!]
    \centering
    \includegraphics[width=\columnwidth]{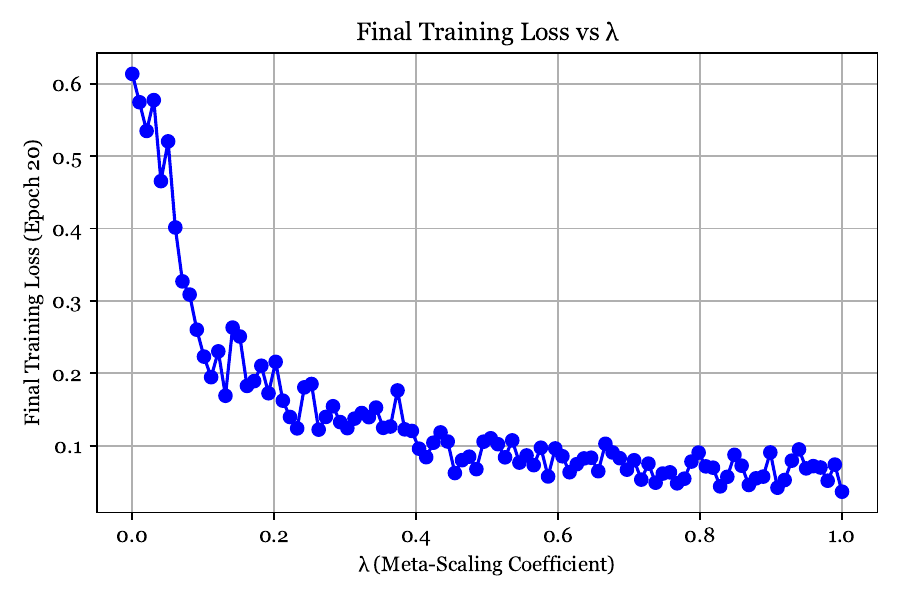}
    \caption{Final Training Loss (Epoch 20) as a function of $\lambda$.}
    \label{fig:final_loss_vs_lambda}
\end{figure}

The \textbf{Final Training Loss vs. $\lambda$} plot (\textit{cf.} Figure~\ref{fig:final_loss_vs_lambda}) clearly demonstrates that increasing $\lambda$ significantly reduces the training loss achieved after 20 epochs. For $\lambda$ values close to zero, the loss remains high (above 0.6), indicating poor optimization. As $\lambda$ increases, especially within the range of $0.0 < \lambda \leq 0.2$, the training loss drops sharply to values below 0.15. This suggests that even a small contribution from the meta-parameters, scaled by $\lambda$, dramatically improves the model's ability to find lower minima in the loss landscape.

\begin{figure}[h!]
    \centering
    \includegraphics[width=\columnwidth]{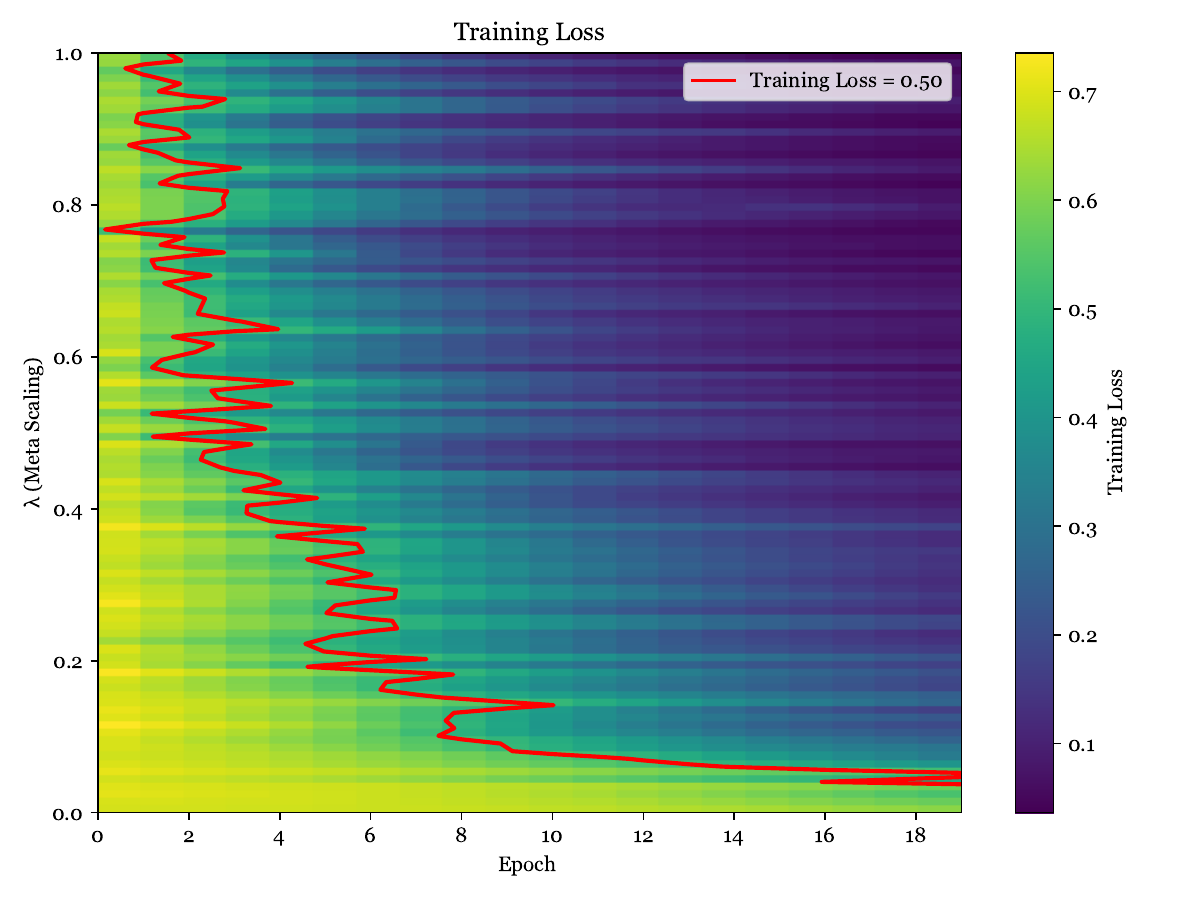}
    \caption{Training Loss heatmap across epochs and $\lambda$ values.}
    \label{fig:heatmap_loss}
\end{figure}

This improvement in loss is corroborated by the \textbf{Training Loss Heatmap} (\textit{cf.} Figure~\ref{fig:heatmap_loss}). This heatmap illustrates the evolution of training loss across epochs for various $\lambda$ values. It vividly shows that for low $\lambda$ (e.g., $\lambda < 0.1$), the training loss converges slowly and remains high, often above 0.5, even after 18 epochs. In contrast, as $\lambda$ increases, the loss quickly drops to lower values within the first few epochs. For $\lambda > 0.2$, the training loss consistently reaches values below 0.1, indicating efficient optimization. The sharp red line, indicating training loss equal to 0.50, prominently descends towards higher epochs as $\lambda$ increases, signifying faster convergence to lower loss values with larger $\lambda$.

\begin{figure}[h!]
    \centering
    \includegraphics[width=\columnwidth]{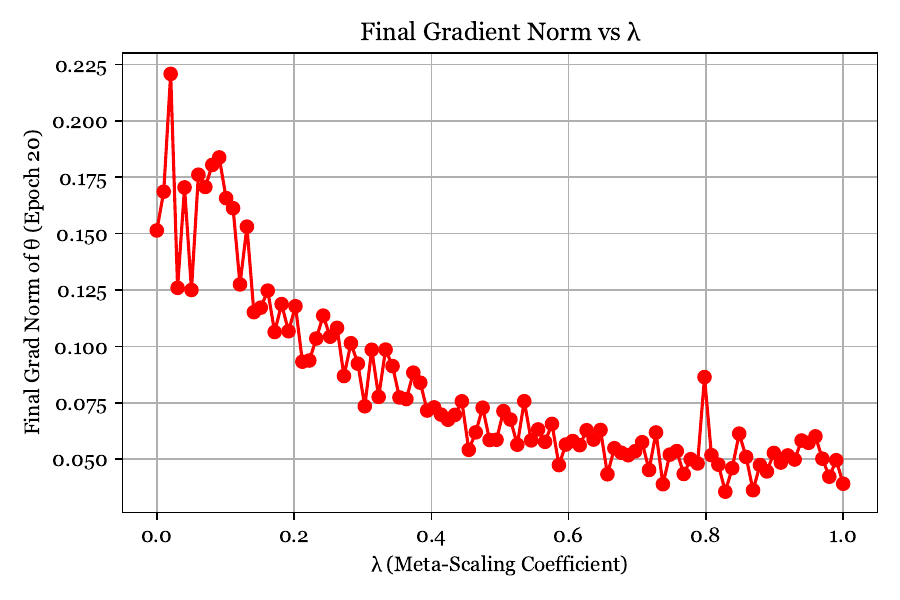}
    \caption{Final Gradient Norm of $\theta$ (Epoch 20) as a function of $\lambda$.}
    \label{fig:final_grad_norm_vs_lambda}
\end{figure}

The \textbf{Final Gradient Norm vs. $\lambda$} plot (\textit{cf.} Figure~\ref{fig:final_grad_norm_vs_lambda}) reveals a strong inverse correlation between $\lambda$ and the final gradient norm of the PQC parameters. For small $\lambda$ values (e.g., $\lambda < 0.1$), the final gradient norm is high, sometimes exceeding 0.2. As $\lambda$ increases, the final gradient norm consistently decreases, stabilizing at values below 0.075 for $\lambda > 0.4$. This trend indicates that higher $\lambda$ values lead to shallower gradients at the end of training, suggesting that the optimization process has more effectively navigated the loss landscape to reach flatter, more stable minima. This reduction in gradient norm implies that the meta-learned parameters contribute to a more isotropic and well-behaved loss landscape, facilitating convergence.

\begin{figure}[h!]
    \centering
    \includegraphics[width=\columnwidth]{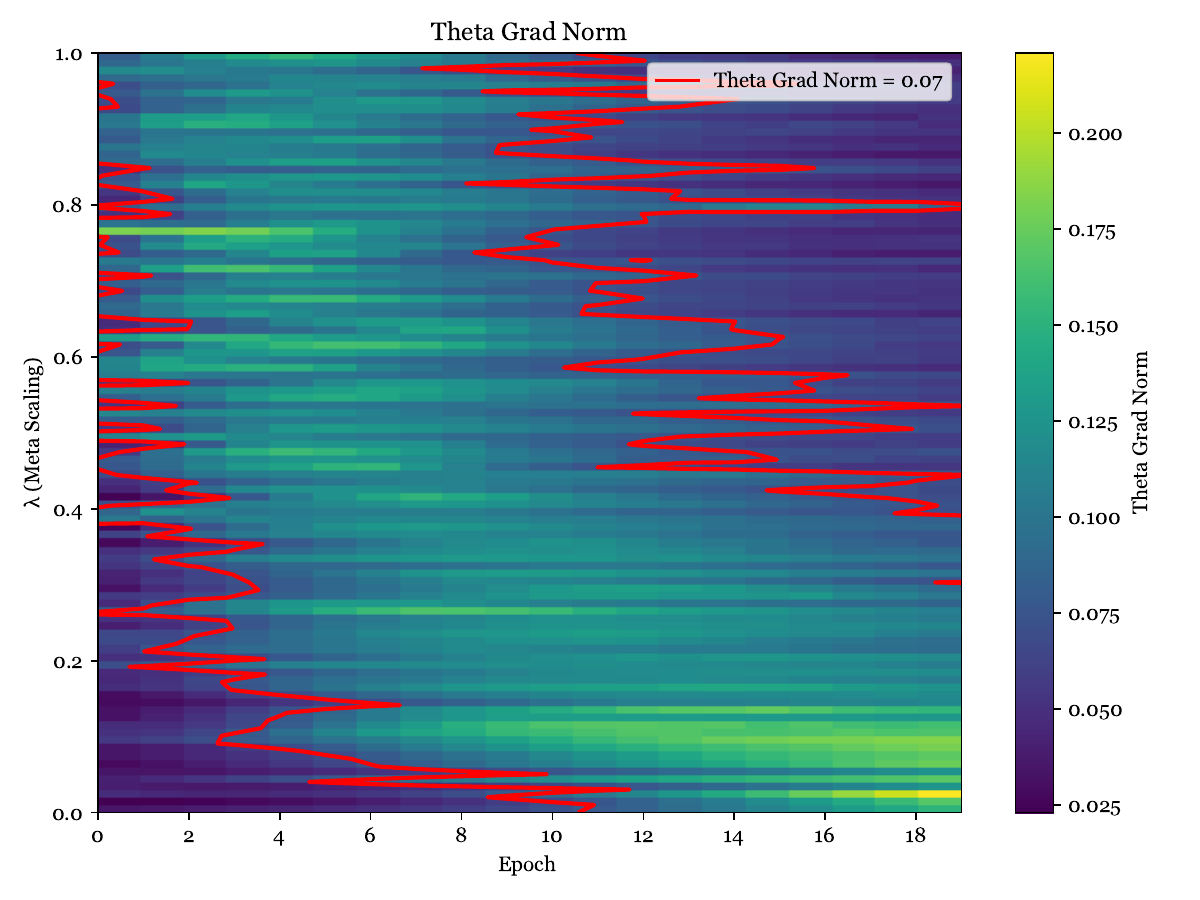}
    \caption{Theta Grad Norm heatmap across epochs and $\lambda$ values.}
    \label{fig:heatmap_grad_norm}
\end{figure}

The \textbf{Theta Grad Norm Heatmap} (\textit{cf.} Figure~\ref{fig:heatmap_grad_norm}) further visualizes this dynamic. For low $\lambda$, the gradient norms remain high throughout training, appearing as lighter shades (yellow) across epochs. Conversely, for larger $\lambda$, particularly above 0.2, the gradient norms quickly diminish to very low values, represented by darker shades (purple). The red line, indicating a gradient norm of 0.07, descends significantly faster for higher $\lambda$ values, signifying that the parameters quickly reach regions where gradients are small. This observation reinforces the idea that increasing the meta-learned contribution regularizes the landscape, making it easier for gradient-based optimizers to find and settle into optimal regions.

\subsection{Generalization Performance: Test Accuracy Dynamics}\label{sec:generalization_performance}

\begin{figure}[h!]
    \centering
    \includegraphics[width=\columnwidth]{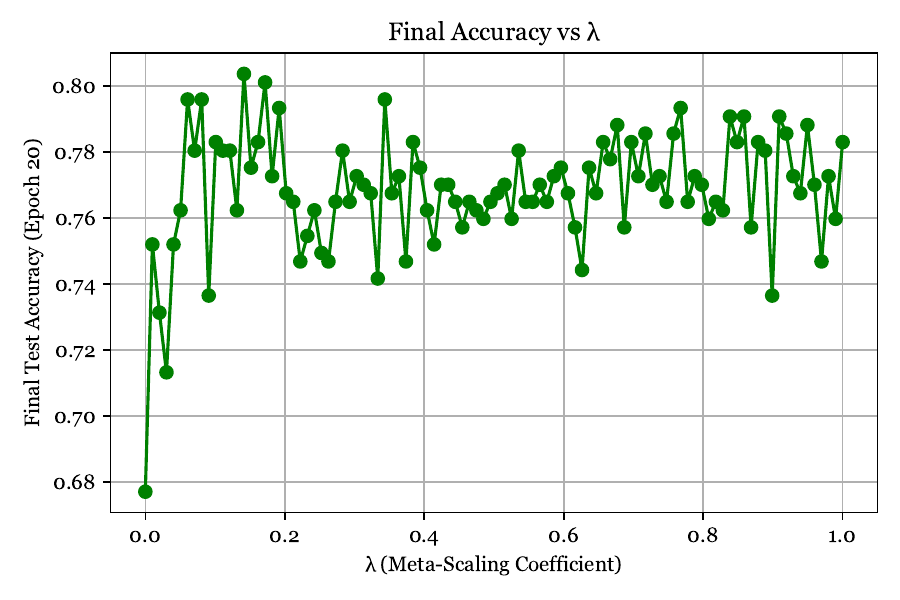}
    \caption{Final Test Accuracy (Epoch 20) as a function of $\lambda$.}
    \label{fig:final_accuracy_vs_lambda}
\end{figure}

The \textbf{Final Test Accuracy vs. $\lambda$} plot (\textit{cf.} Figure~\ref{fig:final_accuracy_vs_lambda}) demonstrates a significant improvement in generalization as $\lambda$ increases. For $\lambda$ values near zero, the final test accuracy is low (around 0.68). A sharp increase in accuracy is observed as $\lambda$ moves from 0.0 to approximately 0.15, where the accuracy often exceeds 0.78 and even reaches above 0.80. Although there are fluctuations, the general trend indicates that the meta-learned parameters (when sufficiently weighted) lead to more robust models that generalize better to unseen data. This supports the theoretical prediction that improved FS metric conditioning acts as a geometric regularizer, enhancing generalization.

\begin{figure}[h!]
    \centering
    \includegraphics[width=\columnwidth]{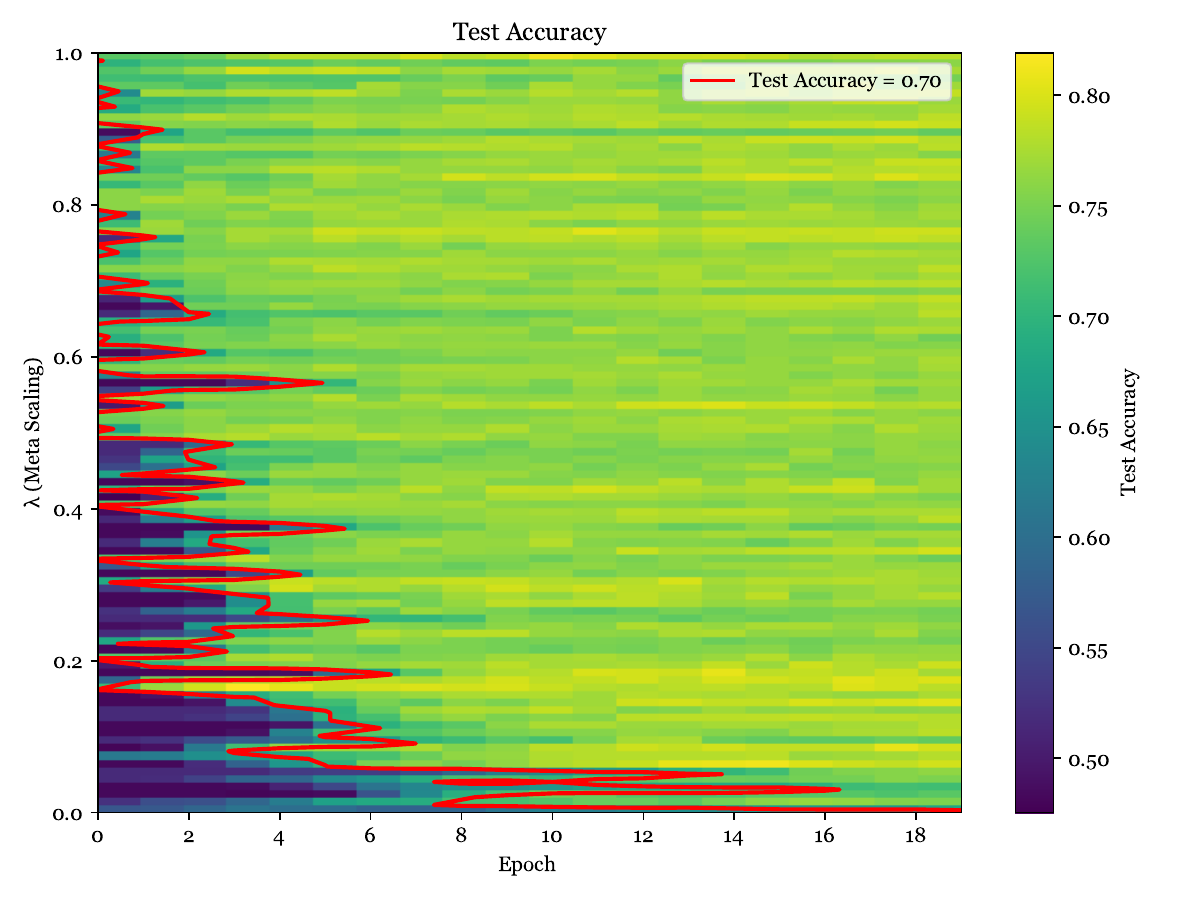}
    \caption{Test Accuracy heatmap across epochs and $\lambda$ values.}
    \label{fig:heatmap_accuracy}
\end{figure}

The \textbf{Test Accuracy Heatmap} (\textit{cf.} Figure~\ref{fig:heatmap_accuracy}) provides a detailed view of test accuracy evolution over epochs for different $\lambda$ values. It clearly shows that for small $\lambda$ values, the test accuracy remains relatively low, often below 0.60, throughout the training process. In contrast, for larger $\lambda$ values (e.g., $\lambda > 0.15$), the model quickly achieves and maintains higher test accuracies, reaching values above 0.75 within a few epochs. The red line, representing test accuracy of 0.70, shifts dramatically towards earlier epochs as $\lambda$ increases, demonstrating that models with higher meta-parameter influence achieve good generalization much faster. This consistent improvement in test accuracy with increasing $\lambda$ underscores the effectiveness of FS metric conditioning in producing models that not only train more efficiently but also generalize more effectively to new data.

\section{Discussion}

This work introduces a novel meta-learning approach for conditioning the Fubini--Study (FS) metric tensor in Parameterized Quantum Circuits (PQCs), demonstrating its profound impact on trainability and generalization in Variational Quantum Algorithms (VQAs). Our theoretical analysis established that the spectral properties of the FS metric, particularly its logarithmic condition number ($\log \kappa$), are pivotal for efficient optimization and tighter PAC-Bayes generalization bounds. The empirical results strongly support these predictions.

The meta-training phase successfully demonstrated that our \textsc{Sculpture} model can effectively learn to generate PQC parameters that lead to a significantly lower logarithmic condition number ($\log \kappa$) of the FS metric. The rapid decrease of $\log \kappa$ from approximately 1.4737 to around 0.6375, accompanied by an increase in $\lambda_{\min}$ and a slight decrease in $\lambda_{\max}$, clearly indicates the meta-learner's ability to promote a more isotropic and well-conditioned parameter space. This geometric regularization directly mitigates the challenge of barren plateaus by ensuring that parameter sensitivities are uniformly distributed, preventing vanishing gradients in certain directions. The low $\log \kappa$ on unseen test data further highlights the meta-learner's robust generalization capability in producing well-conditioned initializations across diverse inputs.

The downstream classification task on the Kaggle diabetes dataset provided compelling evidence for the practical benefits of this geometry-aware learning. Increasing the meta-scaling coefficient $\lambda$ consistently led to a substantial reduction in final training loss and a significant improvement in final test accuracy. Models with higher $\lambda$ values converged faster to lower loss minima and achieved better generalization performance, supporting our hypothesis that FS metric conditioning acts as a powerful geometric regularizer. The observed decrease in gradient norms with increasing $\lambda$ also corroborates the improved landscape, facilitating smoother and more efficient optimization.

The empirical observations reveal a nuanced interplay between the meta-scaling coefficient $\lambda$ and the Fubini--Study (FS) metric's spectral conditioning. While the meta-model is explicitly trained to minimize the logarithmic condition number ($\log \kappa(G)$), aiming to establish a globally more tractable parameter landscape, the application of the composite parameterization $\bm{\theta}(x) = \bm{\theta}_{\mathrm{task}} + \lambda \cdot \bm{\theta}_{\mathrm{meta}}(x)$ in the downstream task introduces a dynamic, data-dependent influence. Here, $\bm{\theta}_{\mathrm{meta}}(x)$, scaled by $\lambda$, acts as an adaptive geometric regularizer, guiding the PQC's parameters towards locally more favorable regions within the loss landscape at each training step for every data instance. Consequently, while a monotonic decrease in the global $\log \kappa(G)$ at a fixed evaluation epoch (e.g., Epoch 20) is not consistently observed across all $\lambda$ values, the overall effect of increasing $\lambda$ is to facilitate a more efficient and stable optimization trajectory. This is evidenced by the consistent reduction in training loss, improved test accuracy, and notably, the decrease in final gradient norms, which signifies a smoother and more navigable landscape where gradients are well-behaved. Thus, the meta-model ensures that the effective landscape encountered by the optimizer during training is more amenable to learning, enabling the model to converge to better solutions and generalize more effectively, even if the instantaneous $\log \kappa(G)$ exhibits some fluctuations.

These findings collectively suggest that sculpting the quantum landscape via meta-learning provides a principled avenue to enhance VQA performance.

\section{Limitations and Future Directions}

Our current study provides compelling evidence for the benefits of FS metric conditioning via meta-learning. However, like any pioneering work, it is subject to certain limitations that open avenues for future research.

\subsection{Limitations}
Firstly, our empirical validation was conducted on a single Parameterized Quantum Circuit (PQC) architecture with a fixed depth and number of qubits. While this allowed for a focused analysis of the meta-learning approach, the generalizability of our findings to other PQC architectures (e.g., hardware-efficient ansatzes, specific chemistry-inspired circuits) remains an open question. Similarly, the scalability of our method to significantly deeper circuits or a larger number of qubits---regimes where barren plateaus become more pronounced---warrants further investigation. Secondly, the downstream task was limited to a specific classical classification problem on a single dataset. Future work should explore the efficacy of FS metric conditioning across a broader range of datasets and problem types, including diverse classical datasets, quantum chemistry simulations, or combinatorial optimization problems, to ascertain its robustness and universal applicability.

\subsection{Future Directions}
Building upon the promising results presented, several exciting directions emerge for future research. A natural extension involves applying our meta-learning framework to more complex and larger-scale quantum tasks, such as those encountered in quantum chemistry, materials science, or advanced machine learning applications, to rigorously test its scalability and performance benefits. Furthermore, to facilitate broader adoption and impact within the quantum computing community, an important step would be to develop a user-friendly Python package that encapsulates the meta-training pipeline and the $\lambda$-scaling mechanism. Ultimately, a significant endeavor would be to train and release a pre-computed, ready-to-use meta-model. Such a meta-model, trained on a diverse suite of PQC architectures and problem types, could potentially provide off-the-shelf geometrically conditioned parameters, significantly lowering the barrier to entry for researchers and practitioners aiming to improve the trainability and generalization of their variational quantum algorithms without the need for extensive meta-training.

\begin{acknowledgments}
\textbf{Author Contributions:} Marwan Ait Haddou was responsible for the conceptualization, methodology, software development, formal analysis, investigation, data curation, and writing of the original draft. Mohamed Bennai provided supervision, theoretical insights, validation, and critical review and editing of the manuscript. Both authors have read and agreed to the published version of the manuscript.

\textbf{Funding:} This research received no specific grant from any funding agency in the public, commercial, or not-for-profit sectors.

\textbf{Conflicts of Interest:} The authors declare no conflicts of interest.

\textbf{Generative AI Usage:} Generative AI tools were employed solely for linguistic enhancement, grammar correction, and typesetting suggestions during the manuscript preparation. The intellectual content, scientific methodology, experimental design, data analysis, and interpretation of results remain the sole responsibility and original work of the listed authors.

\textbf{Data and Code Availability:} All datasets utilized in this study are publicly available. The code for reproducing the experiments is available from the corresponding author upon reasonable request.
\end{acknowledgments}
\newpage
\bibliographystyle{plainnat}
\bibliography{refs}

\newpage
\clearpage 
\begin{onecolumngrid}
\appendix 
\section{Fubini--Study Metric Conditioning in Parameterized Quantum Circuits}

\subsection{Introduction}
The \textbf{Fubini--Study (FS) metric tensor} characterizes the intrinsic geometry of quantum state manifolds induced by parameterized quantum circuits (PQCs). Understanding this metric is crucial for analyzing the trainability, convergence, and generalization of variational quantum algorithms (VQAs). In this appendix, we rigorously derive the FS metric, analyze its spectral structure, and interpret its condition number $\kappa(G)$ from both geometric and information-theoretic standpoints.

We specifically focus on PQCs with a \textbf{composite parameterization}:
\[
\bm{\theta}(x) = \bm{\theta}_{\mathrm{task}} + \lambda \cdot \bm{\theta}_{\mathrm{meta}}(x),
\]
where $\bm{\theta}_{\mathrm{meta}}(x)$ is data-dependent and generated by a frozen meta-model trained to minimize $\log \kappa(G)$. Our investigation explores how this approach improves both optimization and generalization by conditioning the FS metric.

\subsection{The Fubini--Study Metric Tensor: Definitions and Derivations}
Let $\ket{\psi(\bm{\theta})} = U(\bm{\theta}) \ket{\psi_0}$ represent a smooth family of normalized pure states generated by a parameterized unitary PQC, where $\bm{\theta} \in \mathbb{R}^d$ is the vector of parameters.

\paragraph{Intrinsic Geometry and the Fubini--Study Metric}
The \emph{Fubini--Study (FS) metric} quantifies the infinitesimal distinguishability between nearby quantum states on the projective Hilbert space $\mathcal{P}(\mathcal{H})$, which is the space of rays (normalized states modulo global phase).

The infinitesimal squared distance between $\ket{\psi(\bm{\theta})}$ and $\ket{\psi(\bm{\theta} + \dd \bm{\theta})}$ is defined via the \emph{quantum fidelity} as:
\begin{equation}
\dd s^2 := 1 - \abs{\langle \psi(\bm{\theta}) | \psi(\bm{\theta} + \dd \bm{\theta}) \rangle}^2.
\label{eq:fs_distance}
\end{equation}
Let's expand the fidelity term $\langle \psi(\bm{\theta}) | \psi(\bm{\theta} + \dd \bm{\theta}) \rangle$ using a Taylor series around $\bm{\theta}$:
\begin{align*}
\langle \psi(\bm{\theta}) | \psi(\bm{\theta} + \dd \bm{\theta}) \rangle &= \langle \psi | \left( \ket{\psi} + \sum_i \frac{\partial \ket{\psi}}{\partial \theta_i} \dd \theta_i + \frac{1}{2} \sum_{i,j} \frac{\partial^2 \ket{\psi}}{\partial \theta_i \partial \theta_j} \dd \theta_i \dd \theta_j + \mathcal{O}(\|\dd \bm{\theta}\|^3) \right) \\
&= \langle \psi | \psi \rangle + \sum_i \langle \psi | \partial_i \psi \rangle \dd \theta_i + \frac{1}{2} \sum_{i,j} \langle \psi | \partial_{ij}^2 \psi \rangle \dd \theta_i \dd \theta_j + \mathcal{O}(\|\dd \bm{\theta}\|^3),
\end{align*}
where $\ket{\partial_i \psi} := \frac{\partial}{\partial \theta_i} \ket{\psi(\bm{\theta})}$ and $\ket{\partial_{ij}^2 \psi} := \frac{\partial^2}{\partial \theta_i \partial \theta_j} \ket{\psi(\bm{\theta})}$. Since $\ket{\psi(\bm{\theta})}$ is normalized, $\langle \psi | \psi \rangle = 1$, and thus $\frac{\partial}{\partial \theta_i} \langle \psi | \psi \rangle = \langle \partial_i \psi | \psi \rangle + \langle \psi | \partial_i \psi \rangle = 0$, which implies $\Re[\langle \psi | \partial_i \psi \rangle] = 0$.

Now consider the squared magnitude of the fidelity:
\begin{align*}
\abs{\langle \psi(\bm{\theta}) | \psi(\bm{\theta} + \dd \bm{\theta}) \rangle}^2 &= \left( \langle \psi | \psi \rangle + \sum_i \langle \psi | \partial_i \psi \rangle \dd \theta_i + \dots \right) \left( \langle \psi | \psi \rangle + \sum_j \langle \partial_j \psi | \psi \rangle \dd \theta_j + \dots \right) \\
&= 1 + \sum_i (\langle \psi | \partial_i \psi \rangle + \langle \partial_i \psi | \psi \rangle) \dd \theta_i + \sum_{i,j} \left( \langle \psi | \partial_i \psi \rangle \langle \partial_j \psi | \psi \rangle + \frac{1}{2} \langle \psi | \partial_{ij}^2 \psi \rangle + \frac{1}{2} \langle \partial_{ij}^2 \psi | \psi \rangle \right) \dd \theta_i \dd \theta_j + \mathcal{O}(\|\dd \bm{\theta}\|^3).
\end{align*}
Using $\langle \partial_i \psi | \psi \rangle = -\langle \psi | \partial_i \psi \rangle$, the first-order term vanishes. For the second-order term, we utilize the identity $\langle \partial_{ij}^2 \psi | \psi \rangle + \langle \psi | \partial_{ij}^2 \psi \rangle = -\left( \langle \partial_i \psi | \partial_j \psi \rangle + \langle \partial_j \psi | \partial_i \psi \rangle \right)$.
Thus,
\begin{align}
\abs{\langle \psi(\bm{\theta}) | \psi(\bm{\theta} + \dd \bm{\theta}) \rangle}^2 &= 1 + \sum_{i,j} \left( \langle \psi | \partial_i \psi \rangle \langle \partial_j \psi | \psi \rangle - \Re[\langle \partial_i \psi | \partial_j \psi \rangle] \right) \dd \theta_i \dd \theta_j + \mathcal{O}(\|\dd \bm{\theta}\|^3) \nonumber \\
&= 1 - \sum_{i,j=1}^d \Re\left[ \langle \partial_i \psi | \partial_j \psi \rangle - \langle \partial_i \psi | \psi \rangle \langle \psi | \partial_j \psi \rangle \right] \dd \theta_i \dd \theta_j + \mathcal{O}(\|\dd \bm{\theta}\|^3).
\label{eq:fidelity_expansion}
\end{align}
Comparing \eqref{eq:fs_distance} and \eqref{eq:fidelity_expansion}, the \textbf{FS metric tensor} $G(\bm{\theta}) \in \mathbb{R}^{d \times d}$ is
\begin{equation} \label{eq:fs_metric_def}
G_{ij}(\bm{\theta}) = \Re\left[ \langle \partial_i \psi | \partial_j \psi \rangle - \langle \partial_i \psi | \psi \rangle \langle \psi | \partial_j \psi \rangle \right].
\end{equation}
Here, $\ket{\partial_i \psi} := \frac{\partial}{\partial \theta_i} \ket{\psi(\bm{\theta})}$ and $\Re$ denotes the real part.

\textbf{Interpretation:} The FS metric measures the \emph{local distinguishability} of quantum states under infinitesimal parameter shifts, effectively defining a Riemannian metric on the manifold $\mathcal{M} = \{ \ket{\psi(\bm{\theta})} : \bm{\theta} \in \mathbb{R}^d \}$. It coincides with the quantum Fisher information metric for pure states \citep{abbas2021power}.

\paragraph{Representation via Generators and Covariance}
Assume the PQC is composed as
\begin{equation} \label{eq:unitary_prod}
U(\bm{\theta}) = \prod_{k=1}^L \exp\left(-i \frac{\theta_k}{2} \tilde{H}_k \right),
\end{equation}
where each $\tilde{H}_k$ is a Hermitian generator.

The derivative of the state $\ket{\psi(\bm{\theta})} = U(\bm{\theta}) \ket{\psi_0}$ with respect to $\theta_i$ is given by:
\begin{equation} \label{eq:partial_psi}
\ket{\partial_i \psi} = \frac{\partial U(\bm{\theta})}{\partial \theta_i} \ket{\psi_0} = -\frac{i}{2} \left( \prod_{k=1}^{i-1} e^{-i\frac{\theta_k}{2}\tilde{H}_k} \right) \tilde{H}_i \left( \prod_{k=i+1}^{L} e^{-i\frac{\theta_k}{2}\tilde{H}_k} \right) \ket{\psi_0}.
\end{equation}
More generally, if $\tilde{H}_i(\bm{\theta})$ is the generator in the Heisenberg picture, i.e., $\tilde{H}_i(\bm{\theta}) := U^\dagger(\bm{\theta}) H_i U(\bm{\theta})$ for a fixed $H_i$, then
\begin{equation} \label{eq:psi_derivative_heisenberg}
\ket{\partial_i \psi} = -\frac{i}{2} H_i \ket{\psi(\bm{\theta})}.
\end{equation}
For a general PQC where $\frac{\partial U}{\partial \theta_i} = -i J_i U$ for some operator $J_i$, then $\ket{\partial_i \psi} = -i J_i \ket{\psi}$.
Substituting these into \eqref{eq:fs_metric_def}:
The first term is:
\begin{equation} \label{eq:part_psi_sq_fidelity}
\langle \partial_i \psi | \partial_j \psi \rangle = \left(-\frac{i}{2}\right) \left(\frac{i}{2}\right) \langle H_i \psi | H_j \psi \rangle = \frac{1}{4} \langle \psi | H_i H_j | \psi \rangle = \frac{1}{4} \langle H_i H_j \rangle_\psi.
\end{equation}
The second term involves:
\begin{equation} \label{eq:psi_partial_fidelity}
\langle \partial_i \psi | \psi \rangle = \left(-\frac{i}{2}\right) \langle H_i \psi | \psi \rangle = -\frac{i}{2} \langle H_i \rangle_\psi.
\end{equation}
And similarly, $\langle \psi | \partial_j \psi \rangle = \frac{i}{2} \langle H_j \rangle_\psi$.
Substituting \eqref{eq:part_psi_sq_fidelity} and \eqref{eq:psi_partial_fidelity} into \eqref{eq:fs_metric_def}:
\begin{align} \label{eq:fs_metric_deriv}
G_{ij}(\bm{\theta}) &= \Re\left[ \frac{1}{4} \langle H_i H_j \rangle_\psi - \left(-\frac{i}{2} \langle H_i \rangle_\psi\right) \left(\frac{i}{2} \langle H_j \rangle_\psi\right) \right] \nonumber \\
&= \Re\left[ \frac{1}{4} \langle H_i H_j \rangle_\psi - \frac{i^2}{4} \langle H_i \rangle_\psi \langle H_j \rangle_\psi \right] \nonumber \\
&= \Re\left[ \frac{1}{4} \langle H_i H_j \rangle_\psi + \frac{1}{4} \langle H_i \rangle_\psi \langle H_j \rangle_\psi \right] \nonumber \\
&= \frac{1}{4} \left( \langle H_i H_j \rangle_\psi - \langle H_i \rangle_\psi \langle H_j \rangle_\psi \right),
\end{align}
where the last step uses the property that for Hermitian operators $H_i, H_j$, $\langle H_i H_j \rangle_\psi$ and $\langle H_i \rangle_\psi \langle H_j \rangle_\psi$ are real. This can be written as the covariance:
\begin{equation} \label{eq:fs_metric_cov}
G_{ij}(\bm{\theta}) = \frac{1}{4} \mathrm{Cov}_\psi(H_i, H_j).
\end{equation}

\textbf{Remarks:}
\begin{itemize}
\item $G$ is positive semidefinite and symmetric.
\item Small covariance directions correspond to parameters inducing little distinguishability change (potential barren plateaus).
\item The spectrum of $G$ encodes anisotropy in parameter space.
\end{itemize}

\subsection{Spectral Properties and Conditioning} \label{app:spectral_stats}
Let the eigenvalues of the FS metric tensor $G(\bm{\theta})$ be $\lambda_1 \geq \cdots \geq \lambda_d > 0$.

\paragraph{Condition number:} \label{eq:kappa_G}
The condition number $\kappa(G)$ is a standard measure of the numerical stability of a matrix and is defined as the ratio of its largest to its smallest eigenvalue:
\begin{equation} \label{eq:condition_number}
\kappa(G) := \frac{\lambda_{\max}}{\lambda_{\min}} = \frac{\lambda_1}{\lambda_d}.
\end{equation}
A large $\kappa(G)$ implies that the matrix $G$ is ill-conditioned, which can cause unstable and slow convergence in optimization algorithms.

\paragraph{Bounds via diagonal variances:}
A simple bound on the condition number can be established using the diagonal elements, which correspond to the variances of individual generators:
\begin{equation} \label{eq:kappa_bound}
\kappa(G) \geq \frac{\max_i G_{ii}}{\min_i G_{ii}} = \frac{\max_i \mathrm{Var}_\psi(H_i)}{\min_i \mathrm{Var}_\psi(H_i)}.
\end{equation}

\paragraph{Spectral entropy and effective dimension:}
To quantify the isotropy of the parameter space, we can use concepts from information theory. Define normalized eigenvalues $p_i = \frac{\lambda_i}{\sum_j \lambda_j}$.
The spectral entropy $\mathcal{H}(G)$ is then:
\begin{equation} \label{eq:spectral_entropy}
\mathcal{H}(G) := - \sum_i p_i \log p_i.
\end{equation}
And the effective dimension $d_{\mathrm{eff}}$ is:
\begin{equation} \label{eq:effective_dimension}
d_{\mathrm{eff}} := \frac{(\sum_i \lambda_i)^2}{\sum_i \lambda_i^2} = \frac{1}{\sum_i p_i^2}.
\end{equation}
Maximal isotropy in the parameter space, where all eigenvalues are equal ($\lambda_i = \text{constant}$), implies $\mathcal{H}(G) \to \log d$ and $d_{\mathrm{eff}} \to d$.

\paragraph{Volume element and model complexity:} \label{app:volume_element}
The volume element of the parameter space, with respect to the FS metric, is given by the square root of the determinant of $G$:
\begin{equation} \label{eq:volume_element}
\mathrm{Vol}(G) := \sqrt{\det G} = \sqrt{\prod_i \lambda_i}.
\end{equation}
This quantity reflects the effective capacity or complexity of the model. Flattening the spectrum (making eigenvalues more uniform) increases this volume and has been shown to reduce overfitting \citep{amari2016information}.

\subsection{Optimization and Natural Gradient Flow} \label{app:optimization_analysis}
Consider a general objective function $f(\bm{\theta})$. Standard gradient descent updates parameters as:
\begin{equation} \label{eq:gd_update}
\bm{\theta}_{k+1} = \bm{\theta}_k - \eta \nabla f(\bm{\theta}_k).
\end{equation}

\paragraph{Natural gradient:}
The natural gradient is derived from performing the steepest descent in the Riemannian manifold defined by the FS metric. The update rule is given by:
\begin{equation} \label{eq:nat_grad_update}
\bm{\theta}_{k+1} = \bm{\theta}_k - \eta G(\bm{\theta}_k)^{-1} \nabla f(\bm{\theta}_k).
\end{equation}
This update preconditions the gradient by the inverse of the metric tensor, effectively transforming the Euclidean gradient into a direction that corresponds to the steepest descent in the intrinsic geometry of the state space.

Bounds on natural gradient norm:
The norm of the natural gradient is bounded by the eigenvalues of $G$. For any gradient vector $\nabla f$, the natural gradient norm $\|G^{-1} \nabla f\|_2$ satisfies:
\begin{equation} \label{eq:nat_grad_norm_bounds}
\frac{1}{\lambda_{\max}} \|\nabla f\|_2 \leq \| G^{-1} \nabla f \|_2 \leq \frac{1}{\lambda_{\min}} \|\nabla f\|_2.
\end{equation}
This highlights that when $\lambda_{\min}$ is very small (leading to a large $\kappa$), the natural gradient norm can become extremely large, potentially causing instability, although its \textit{direction} is optimal. Conversely, if $\lambda_{\max}$ is very large, the natural gradient can become small. The key benefit is that the natural gradient rescales the learning rate differently along various directions, preventing vanishing gradients in flat directions.

\paragraph{Convergence rates:}
For a quadratic objective function $f(\bm{\theta}) = \frac{1}{2} (\bm{\theta} - \bm{\theta}^*)^T A (\bm{\theta} - \bm{\theta}^*)$, where $A$ is a positive definite matrix, the convergence rate of standard gradient descent is heavily dependent on the condition number of $A$. When the objective function is effectively defined by the FS metric itself, i.e., $A=G$, the convergence factor for standard gradient descent is given by:
\begin{equation} \label{eq:grad_descent_convergence}
\rho^2 = \left(\frac{\kappa(G) - 1}{\kappa(G) + 1}\right)^2,
\end{equation}
where $\kappa(G)$ is the condition number of the FS metric. This equation clearly demonstrates that a smaller $\kappa(G)$ (i.e., better conditioning) leads to a smaller $\rho^2$, implying faster linear convergence of the optimization process. Notably, for the quadratic objective, natural gradient descent converges in one step if $\eta=1$.

\subsection{Meta-Learning and Conditioning via \texorpdfstring{$\lambda$}{lambda}-Scaling} \label{app:meta_model_training}

Our approach employs a composite parameterization:
\begin{equation} \label{eq:composite_param}
\bm{\theta}(x) = \bm{\theta}_{\mathrm{task}} + \lambda \cdot \bm{\theta}_{\mathrm{meta}}(x),
\end{equation}
where $\bm{\theta}_{\mathrm{meta}}(x)$ are data-dependent parameters generated by a meta-model trained to minimize $\log \kappa(G)$. The FS metric for this parameterized circuit is then 
\[
G_\lambda(x) := G(\bm{\theta}_{\mathrm{task}} + \lambda \bm{\theta}_{\mathrm{meta}}(x),\, x).
\]

\textbf{Locality of conditioning.} Since the FS metric depends on the specific parameter point and the input data, the condition number $\log \kappa(G_\lambda(x))$ is inherently a \emph{local quantity}, evaluated per datapoint $x$. Therefore, to rigorously analyze how the $\lambda$-scaling affects the spectral conditioning, we must consider each $x$ individually. That is, $\log \kappa(G_\lambda)$ is not a global property of the circuit, but a data- and parameter-dependent measure that varies across the dataset and optimization trajectory. Any geometric benefit obtained by modulating $\lambda$ arises from improvements in the \emph{local conditioning} at each $\bm{\theta}(x)$.

The derivative of the logarithmic condition number with respect to $\lambda$ for a fixed input $x$ is given by:
\begin{equation} \label{eq:d_log_kappa_d_lambda}
\begin{split}
\frac{d}{d \lambda} \log \kappa(G_\lambda(x)) 
&= \frac{d}{d \lambda} \left( \log \lambda_{\max} - \log \lambda_{\min} \right) \\
&= \frac{1}{\lambda_{\max}} \frac{d \lambda_{\max}}{d \lambda} 
  - \frac{1}{\lambda_{\min}} \frac{d \lambda_{\min}}{d \lambda}.
\end{split}
\end{equation}

The derivatives of the eigenvalues with respect to $\lambda$ can be computed using standard eigenvalue perturbation theory. For a simple eigenvalue $\lambda_i$ of $G_\lambda(x)$ with corresponding eigenvector $\ket{v_i}$, we have:
\begin{equation} \label{eq:d_lambda_i_d_lambda}
\frac{d \lambda_i}{d \lambda} = \langle v_i | \frac{d G_\lambda(x)}{d \lambda} | v_i \rangle.
\end{equation}

This shows how $\lambda$-scaling directly alters the FS spectrum \emph{locally}, through its influence on the parameter trajectory $\bm{\theta}(x)$. Therefore, understanding the effectiveness of meta-learning and geometric regularization via $\lambda$ requires examining the behavior of $\log \kappa(G_\lambda(x))$ across individual datapoints, and potentially averaging over the input distribution to estimate global trends.

\subsection{Surrogate PAC-Bayes Generalization Bound} \label{app:pac_bayes_derivation}

We derive a PAC-Bayes–inspired generalization bound tailored to parameterized quantum circuits (PQCs), where the geometry of the quantum state manifold is characterized by the Fubini--Study (FS) metric tensor $G(\bm{\theta})$. Our objective is to show that, under natural assumptions, the generalization gap can be upper bounded in terms of the spectral properties of $G$, specifically via the surrogate quantity $\Tr(G)/\lambda_{\min}(G)$.

\paragraph{Classical PAC-Bayes Framework.}
Let $P$ be a prior distribution over parameters $\bm{\theta}$, and $Q$ be a posterior distribution after observing $N$ training examples. Then, for bounded loss functions $\ell \in [0,1]$, the standard PAC-Bayes generalization bound (e.g., \cite{mcallester1999pac, dziugaite2017computing}) states that, with probability at least $1 - \delta$ over training data, the expected (true) risk is bounded by:
\begin{equation}
\mathcal{R}(Q) \leq \hat{\mathcal{R}}_N(Q) + \sqrt{ \frac{1}{2N} \left( KL(Q \| P) + \log\frac{2\sqrt{N}}{\delta} \right) },
\end{equation}
where $\hat{\mathcal{R}}_N(Q)$ is the empirical risk and $KL(Q \| P)$ is the Kullback–Leibler divergence between $Q$ and $P$.

\paragraph{Gaussian Posterior and Geometry.}
To make this bound tractable in our setting, we assume:
\begin{itemize}
    \item The posterior $Q = \mathcal{N}(\bm{\theta}, \Sigma_Q)$ is a multivariate Gaussian centered at a fixed parameter vector $\bm{\theta}$,
    \item The prior $P = \mathcal{N}(0, \Sigma_P)$ is also Gaussian,
    \item The posterior covariance is aligned with the inverse FS metric: $\Sigma_Q \propto G(\bm{\theta})^{-1}$.
\end{itemize}
This last assumption is geometrically motivated by natural gradient descent, where the FS metric defines the local curvature of the loss landscape \cite{amari1998natural}.

Under these assumptions, the KL divergence between $Q$ and $P$ becomes:
\begin{equation}
KL(Q \| P) = \frac{1}{2} \left[ \Tr(\Sigma_P^{-1} \Sigma_Q) + (\bm{\theta} - \bm{\theta}_0)^\top \Sigma_P^{-1} (\bm{\theta} - \bm{\theta}_0) - p + \log \frac{\det \Sigma_P}{\det \Sigma_Q} \right].
\end{equation}
Letting $\Sigma_P = \sigma^2 I$ and $\Sigma_Q = \alpha G^{-1}$, we obtain:
\begin{equation}
KL(Q \| P) = \frac{1}{2} \left[ \frac{\alpha}{\sigma^2} \Tr(G^{-1}) + \frac{1}{\sigma^2} \|\bm{\theta} - \bm{\theta}_0\|^2 - p + \log \frac{\sigma^{2p}}{\det(\alpha G^{-1})} \right].
\end{equation}
After simplification and absorbing constants into a global constant $C$, we write:
\begin{equation}
KL(Q \| P) \leq C \left[ Tr(G^{-1}) + \log\det G + \|\bm{\theta}\|^2 \right].
\end{equation}

\paragraph{Spectral Surrogates.}
To further simplify, let $\{\lambda_i\}_{i=1}^p$ be the eigenvalues of $G$. Then:
\[
\Tr(G^{-1}) = \sum_{i=1}^p \frac{1}{\lambda_i}, \quad \log \det G = \sum_{i=1}^p \log \lambda_i.
\]
We upper-bound $\Tr(G^{-1})$ by:
\[
\Tr(G^{-1}) \leq \frac{p}{\lambda_{\min}(G)},
\]
and approximate $\log \det G$ via $\log \Tr(G)$ and Jensen’s inequality. Then:
\begin{equation*}
    KL(Q \| P) \lesssim \frac{p \cdot Tr(G)}{\lambda_{\min}(G)} + \text{(bounded terms)}.
\end{equation*}

\paragraph{Surrogate Generalization Bound.}
Plugging this estimate into the PAC-Bayes inequality gives:
\begin{equation}
\mathcal{R}(\bm{\theta}) \leq \hat{\mathcal{R}}_N(\bm{\theta}) + \sqrt{ \frac{1}{2N} \left( \frac{C \cdot \Tr(G)}{\lambda_{\min}(G)} + \log \frac{1}{\delta} \right) }.
\end{equation}
Absorbing constants and simplifying, we obtain the final surrogate generalization bound:
\begin{equation} \label{eq:fs_pac_bayes_surrogate}
\mathcal{R}(\bm{\theta}) \leq \hat{\mathcal{R}}_N(\bm{\theta}) + \sqrt{ \frac{C}{N} \cdot \frac{\Tr(G)}{\lambda_{\min}(G)} } \cdot \log \frac{1}{\delta}
\end{equation}
where $C$ absorbs dimension-dependent terms and loss regularity constants.

\paragraph{Interpretation.}
This expression is not a formally derived PAC-Bayes bound, but a spectral surrogate that links generalization to the curvature of the quantum model’s geometry. The quantity $\Tr(G)/\lambda_{\min}(G)$ captures the effective anisotropy or ill-conditioning of the quantum state manifold. Minimizing this ratio through meta-learning yields flatter and more isotropic geometry, which in turn improves convergence and generalization.

\paragraph{Related Work.}
While this exact bound does not appear in prior literature, it is motivated by the PAC-Bayes flatness bounds of \citet{dziugaite2017computing} and classical geometry-based generalization theory. It also aligns with recent analyses of quantum Fisher information in the context of variational quantum learning \cite{abbas2021power}.

\subsection{Summary: FS Conditioning as Geometric Regularization}
Minimizing $\log \kappa(G)$:
\begin{enumerate}
\item Promotes isotropy in parameter sensitivity.
\item Maximizes spectral entropy $\mathcal{H}(G) \to \log d$.
\item Maximizes effective rank $d_{\mathrm{eff}} \to d$.
\item Reduces optimization anisotropy and accelerates convergence.
\item Tightens PAC-Bayes generalization bounds.
\end{enumerate}

\begin{table*}[htb!]
\centering
\setlength{\tabcolsep}{6pt}
\begin{tabularx}{\textwidth}{@{} >{\bfseries}l X @{}}
\toprule
\textbf{Quantity} & \textbf{Effect of Minimizing $\log \kappa(G)$} \\
\midrule
Gradient Norms & More isotropic, balanced step sizes across parameters \\
Convergence Rate & Faster linear convergence of gradient-based optimization \\
Spectral Entropy $\mathcal{H}(G)$ & Approaches $\log d$, indicating uniform spectral spread \\
Effective Dimension $d_{\mathrm{eff}}$ & Approaches $d$, indicating full utilization of parameter space \\
Generalization Bound & Tighter PAC-Bayes bound with lower complexity \\
\bottomrule
\end{tabularx}
\caption{Summary of the impacts of FS metric conditioning on quantum learning dynamics and generalization.}
\label{tab:fs_conditioning_summary}
\end{table*}

\subsection{Conclusion}
The Fubini--Study metric encapsulates essential geometry for variational quantum algorithms. Conditioning it, especially minimizing its log condition number, yields faster convergence, more uniform parameter sensitivity, and improved generalization. Meta-learning architectures leveraging data-dependent parameterizations and $\lambda$-scaling provide an effective mechanism for achieving this goal. The FS metric thus bridges optimization theory, differential geometry, and statistical learning in the quantum regime.
\end{onecolumngrid}
\end{document}